\pdfoutput=1
\documentclass[11pt]{article}
\usepackage[]{acl}

\usepackage{times}
\usepackage{latexsym}

\usepackage[T1]{fontenc}
\usepackage[utf8]{inputenc}

\usepackage{microtype}

\usepackage{notoccite}
\usepackage{subcaption}
\usepackage{amsmath}
\usepackage{amsfonts}
\usepackage{booktabs}
\usepackage{tikz-dependency}
\usepackage{verbatimbox}
\usepackage{cleveref}
\crefformat{section}{\S#2#1#3}
\crefformat{subsection}{\S#2#1#3}
\crefformat{subsubsection}{\S#2#1#3}
\crefmultiformat{section}{\S#2#1#3}{ and~\S#2#1#3}{, \S#2#1#3}{, and~\S#2#1#3}
\crefmultiformat{subsection}{\S#2#1#3}{ and~\S#2#1#3}{, \S#2#1#3}{, and~\S#2#1#3}
\crefmultiformat{subsubsection}{\S#2#1#3}{ and~\S#2#1#3}{, \S#2#1#3}{, and~\S#2#1#3}
\crefrangeformat{section}{\mbox{\S#3#1#4--#5#2#6}}
\crefrangeformat{subsection}{\mbox{\S#3#1#4--#5#2#6}}
\crefrangeformat{subsubsection}{\mbox{\S#3#1#4--#5#2#6}}
\crefname{appendix}{Appendix}{Appendices}
\crefname{figure}{Figure}{Figures}
\crefname{subfigure}{Figure}{Figures}
\crefname{table}{Table}{Tables}
\DeclareUnicodeCharacter{0307}{\'{a}}

\title{On the Role of Pre-trained Language Models in Word Ordering:\\ A Case Study with BART}

\author{Zebin Ou\textsuperscript{1}, Meishan Zhang\textsuperscript{3}, Yue Zhang\textsuperscript{1,2}\thanks{\ \ Corresponding Author}\\
  \textsuperscript{1}School of Engineering, Westlake University\\
  \textsuperscript{2}Institute of Advanced Technology, Westlake Institute for Advanced Study \\
  \textsuperscript{3}Institute of Computing and Intelligence, Harbin Institute of Technology (Shenzhen)\\
  \texttt{\{ouzebin,zhangyue\}@westlake.edu.cn}, \\
  \texttt{mason.zms@gmail.com}
}

\begin{document}
\maketitle

\begin{abstract}
    Word ordering is a constrained language generation task taking unordered words as input. Existing work uses linear models and neural networks for the task, yet pre-trained language models have not been studied in word ordering, let alone why they help. We use BART as an instance and show its effectiveness in the task. To explain why BART helps word ordering, we extend analysis with probing and empirically identify that syntactic dependency knowledge in BART is a reliable explanation. We also report performance gains with BART in the related partial tree linearization task, which readily extends our analysis.\footnote{We release our source code at \url{https://github.com/simtony/BART-word-orderer}}
\end{abstract}

\section{Introduction}
\label{sec:introduction}
The task of word ordering \cite{od:wang,od:zhang15,od:tao}, also known as linearization \cite{od:liu}, aims to assign a valid permutation to a bag of words for a coherent sentence. While early work uses word ordering to improve the grammaticality of machine-generated sentences \cite{od:wang}, the task subsequently manifests itself in applications such as discourse generation \cite{od:use:althaus}, machine translation \cite{od:use:eisner,od:use:he}, and image captioning \cite{od:use:fang}. It plays a central role in linguistic realization \cite{gen} of pipeline text generation systems. Advances in word ordering are also relevant to retrieval augmented generation \cite{od:use:guu}, with outputs additionally conditioned on retrieved entries, which can constitute a bag of words.
\begin{figure}[t]
    \centering
    \includegraphics[scale=0.4]{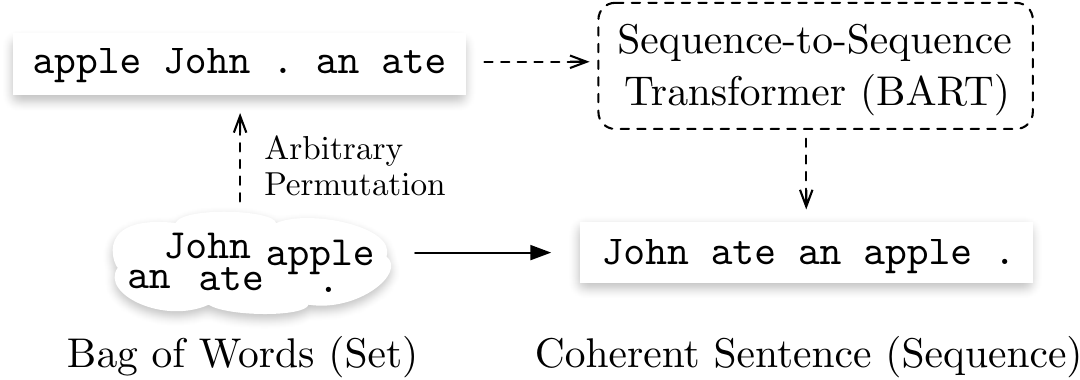}
    \caption{Illustration of our approach for word ordering. Bag-of-words inputs are first turned into pseudo-sequences with arbitrarily word permutations, and then fed to a sequence-to-sequence Transformer for coherent outputs.}
    \label{fig:diagram}
\end{figure}
Word ordering can be viewed as constrained language generation with all inflected output words provided, which makes it more amenable for error analysis (\cref{sec:wo:constrain}). The task can be extended to tree linearization \cite{tree} or partial tree linearization \cite{partial:yue} with syntactic features as additional input. Syntactic models \cite{od:liu} and language models \cite{od:rush} have been used in word ordering to rank candidate word permutations. Recently, \citet{od:hasler} and \citet{od:tao} explore different designs of neural models for the task. However, no existing studies investigate pre-trained language models (PLMs; \citealt{plmsurvey}) for word ordering, which have effectively improved various NLP tasks.

Intuitively, the rich knowledge in PLMs can readily help word ordering. However, the unordered bag-of-words inputs may seem incompatible to PLMs with sequential inputs. The role of PLMs in word ordering thus remains an interesting research question. We fill the research gap by empirically investigating BART \cite{bart}, a pre-trained sequence-to-sequence Transformer \cite{attention}, as an instance of PLMs for word ordering. Specifically, we assign an arbitrary permutation for the input bag of words to obtain a pseudo-sequence, and use sequence-to-sequence Transformers to generate ordered outputs, as illustrated in \cref{fig:diagram}. As BART uses subword inputs \cite{bpe} instead of words \cite{od:rush,od:hasler,od:tao}, we implement constrained beam search using prefix trees (\cref{sec:wo:modeling}; \cref{fig:prefix_horizon}). Results show that BART substantially improves word ordering compared to our Transformer baseline, which already outperforms previous best (\cref{sec:wo:results}).

% Specifically, we tackle the following research questions:

%   \begin{itemize}
%     \item[Q1] Can BART help word ordering?
%     \item[Q2] Why does BART help word ordering?
% \end{itemize}

With Transformer models (including BART), we further investigate consequences of two major modeling decisions, which remain unexamined in the literature. First, while all previous studies assume output sequences constrained within permutations of input words, \citet{od:tao} eliminate such constraints. We find the latter leads to a consistent performance drop, which is further attributed to missing words in outputs, a phenomenon related to the coverage issue in machine translation \cite{coverage}. Second, \citet{od:hasler} use conditional probability $p(\boldsymbol{y}|\boldsymbol{x})$ instead of the unconditional one $p(\boldsymbol{y})$ of \citet{od:rush} to score candidate output permutations $\boldsymbol{y}$ of input bag of words $\boldsymbol{x}$. We provide the first fair comparison for the two approaches, and find that with small decoding beam, conditional models substantially outperform unconditional ones. Interestingly, such an advantage fails to persist as we increase the beam size.

% On the closely related CommonGen task, \citet{commongen} adopt a similar sequential PLM and use data augmentation and mean pooling encoders to address the potential sensitivity.
Our Transformer word orderers may be sensitive to arbitrary word permutations in the input pseudo-sequence (\cref{sec:wo:perm}). Recent studies \cite{perm:sinha,perm:etti} show that Transformers are relatively insensitive to word permutations in \emph{sequential} inputs. They are more sensitive to local orders than global orders of input subwords on the GLUE benchmark \cite{perm:clou}. In contrast, we find that Transformer (including BART) word orderers are relatively insensitive to \emph{both} word and subword permutations in inputs. Such result can be relevant to modeling unordered inputs with PLMs \cite{rdf2text,commongen}.

We finally aim to explain why BART helps word ordering (\cref{sec:why}). Analysis with probing \cite{bertology} provides \emph{speculated} explanations for the utility of PLMs with the possession of numerous types of knowledge. However, for a reliable explanation, we need to identify the \emph{specific} type of knowledge \emph{relevant} to the task. In addition, the amount of the knowledge should be nontrivial \emph{in} the PLM. With a procedure based on feature importance \cite{ablation} and probing \cite{probe:syntax}, we empirically identify that knowledge about syntactic dependency structure reliably explains why BART helps word ordering. Our analysis can be readily extended to partial tree linearization \cite{partial:yue}, for which we also report performance gains with our models (\cref{sec:partial}).

% A reliable explanation depends on the knowledge \emph{relevant} to word ordering. In addition, the amount of the knowledge in BART should be nontrivial. We use a procedure based on feature importance \cite{ablation} and probing \cite{probe:syntax} to find that knowledge about syntactic dependency structure reliably explain why BART helps word ordering.

% With , we
% Although numerous types of knowledge have been identified in PLMs by probing \cite{bertology}, they do not necessarily improve the target task \cite{probe:improve}. A reliable explanation depends on the knowledge \emph{relevant} to word ordering. In addition, the amount of the knowledge in BART should be nontrivial.

% We follow the same approach

% and further identify specifically  \emph{specific} type of knowledge \emph{relevant} to word ordering.

% making such speculations reliable explanations.

\section{Related Work}

\paragraph{Word Ordering Modeling}
Early work uses syntactic models \cite{od:zhangccg,od:liu} and language models \cite{od:zhangccang,od:liu2} to rank candidate permutations of input words. \citet{od:liu2} and \citet{od:rush} discuss their relative importance. Syntactic models rank candidates with the probability of the jointly predicted parse tree. They can be linear models \cite{od:wang} or neural networks \cite{od:song} with hand-crafted features. Language models use the probability of the output sentence for ranking. Early work uses statistical n-gram models \cite{od:zhangccang}, which are later replaced by recurrent neural networks \cite{od:rush}. Most related to our work, \citet{od:hasler} and \citet{od:tao} formulate word ordering as conditional generation. \citet{od:hasler} uses an LSTM decoder with attention \cite{rnnsearch} and an encoder degenerating to an embedding layer. \citet{od:tao} stack self-attention \cite{attention} layers as the encoder and use a decoder from pointer network \cite{pointer}. Both encode bag-of-words inputs with permutation invariant \emph{word} encoders. In contrast, we turn bag-of-words inputs into \emph{subword} pseudo-sequences and feed them to standard sequence-to-sequence models. Instead of investigating features, prediction targets, and model architectures as in previous work, we focus on the utility of BART in the task.

\paragraph{Word Ordering Decoding}
Early work relies on time-constrained best-first-search \cite{od:white,od:zhangccg}. As it lacks an asymptotic upper bound for time complexity \cite{od:liu}, later work with syntactic models \cite{od:song}, language models \cite{od:rush}, and conditional generation models \cite{od:hasler,od:tao} adopt beam search for decoding. All previous work assumes an output space constrained to permutations of input words except for \citet{od:tao}, who assume the output to be any sequences permitted by the vocabulary. However, the effect of such unconstrained output space is unexamined. We compare the difference between beam search with constrained and unconstrained output spaces.

\paragraph{Tasks Related to Word Ordering} Word ordering was first proposed by \citet{od:bang} as a surrogate for grammaticality test, and later formulated by \citet{od:wang} as a standard task. A closely related task is CommonGen \cite{commongen}, which aims to generate a coherent sentence subjective to \emph{commonsense} constraints given a set of \emph{lemmatized} concept words. In contrast, word ordering is a constrained \emph{language modeling} task given \emph{inflected} output words. Tree linearization \cite{tree} is a related task with full dependency trees as inputs. Dropping subsets of dependency arcs and part-of-speech tags results in partial tree linearization \cite{partial:yue}. Further removing functional words and word inflections results in surface realization \cite{surface}. Different from CommonGen and surface realization, the provided output bag-of-words limit reliance on domain knowledge and reduce ambiguity in output, making word ordering a concentrated case for testing generic linguistic capacity \cite{raji2021ai} of text generation models. In addition, word ordering requires no labeling in contrast to all these tasks.

\paragraph{PLMs and Non-Sequential Inputs} PLMs with the Transformer \cite{attention} decoder are amenable for sequence generation \cite{T5,bart}. They have been used for sequence generation tasks with non-sequential inputs, such as AMR-to-Text \cite{penman}, RDF-to-Text \cite{ribeiro-etal-2021-investigating}, and CommonGen \cite{commongen}. Typically, non-sequential inputs are turned into sequential ones before being fed to PLMs. Additionally aiming to understand why BART helps word ordering, we adopt a similar approach and refrain from task-specific engineering, which allows the same sequence-to-sequence model for multiset and tree inputs, limiting extra confounding factors in our analysis.

\paragraph{Analysis with Probing} Previous work on probing \cite{bertology} has identified various types of knowledge in PLMs, such as syntax \cite{probe:syntax}, semantics \cite{probe:semantic}, and commonsense \cite{probe:common}. They are speculated to explain the utility of PLMs in target tasks. We make such explanations reliable for BART in word ordering by establishing the \emph{relevance} of specific types of knowledge to the task, in addition to probing their existence in BART.

\section{Word Ordering with BART}
\label{sec:wo}
We describe our formulation of word ordering and how to adopt the sequence-to-sequence BART for the task (\cref{sec:wo:modeling}), report results on the standard PTB benchmark (\cref{sec:wo:settings,sec:wo:results}), and analyze effects of different modeling decisions (\cref{sec:wo:conditional,sec:wo:constrain,sec:wo:perm}).

\subsection{Modeling Word Ordering}
\label{sec:wo:modeling}
We formulate word ordering as conditional generation following \citet{od:hasler}.
The input bag of words constitutes a multiset $\boldsymbol{x}$, where different elements can take the same value. The probability of output sequence $\boldsymbol{y}$, conditioning on $\boldsymbol{x}$, is parameterized by $\theta$ and factorizes auto-regressively:
\begin{equation}
    p_\theta(\boldsymbol{y} | \boldsymbol{x}) = \prod_{t} p_\theta(y_{t} | \boldsymbol{y}_{<t}, \boldsymbol{x})
\end{equation}
where $\boldsymbol{y}_{<t}$ consists of previous generated tokens up to step $t-1$, the next token $y_t$ takes a token from the output vocabulary. Output sequences start with a special token $y_0$ denoting beginning of sentences. Following \citet{od:hasler}, after solving $\theta$ with maximum likelihood estimation on the training set, we use beam search in an output space for the candidate $\boldsymbol{y}$ maximizing the product $\prod_{t} p_\theta(y_{t} | \boldsymbol{y}_{<t}, \boldsymbol{x})$.

\begin{figure}[t]
    \centering
    \includegraphics[scale=0.3]{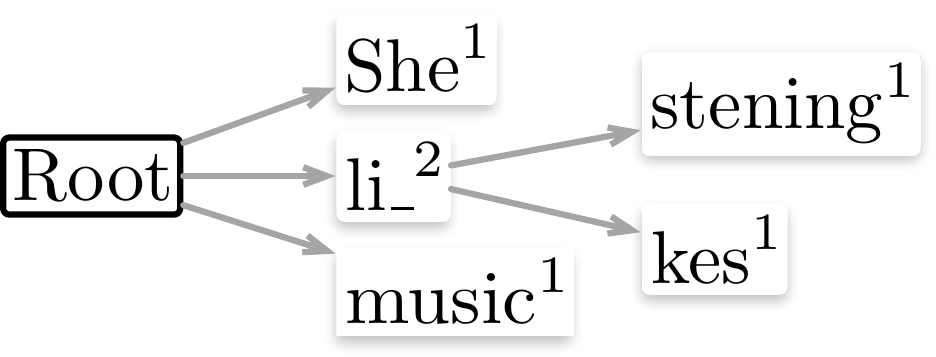}
    \caption{Prefix tree at initialization for constraints \{``She'', ``li\_ kes'', ``li\_ stening'', ``music''\}. Each path from the root to a leaf corresponds to a subword sequence. Except for the root, each node corresponds to a subword and tracks its count (superscript) in the constraints. A pointer (bold outline) denoting the previously generated subword $y_{t-1}$ points to the root at initialization. At decoding step $t$, only subwords of child nodes with nonzero count are valid as the next token $y_t$. Once a subword is selected for $y_t$, we move the pointer to the corresponding node and decrement its count by 1. The pointer is reset to the root after reaching leaves. Decoding for each candidate finishes when all counts are zero.}
    \label{fig:prefix_horizon}
\end{figure}

Output sequences $\boldsymbol{y}$'s are generally constrained within permutations of input words \cite{od:rush,od:tao}. To account for corrupted inputs (e.g., word deletion), \citet{od:tao} use an unconstrained output space with any sequence permitted by the vocabulary. We analyze the output difference between decoding with constrained and unconstrained output space in \cref{sec:wo:constrain}.

Beam search with bag-of-words constraints can be simply implemented by tracking words to be generated using a multiset for each candidate in the beam and setting $p_\theta(y_{t} | \boldsymbol{y}_{<t}, \boldsymbol{x})$ of invalid (not in the multiset) next words to zero. The generated word $y_t$ are then removed from the multiset after decoding step $t$. Decoding of each candidate ends with an empty multiset. Different from previous work \cite{od:rush,od:hasler,od:tao}, subword segmentation of BART turns each input word into a sequence of subwords, which poses sequential constraints on the output. Thus tracking generated subwords and eliminate invalid next subwords during beam search require a different data structure. We compile subword sequences into a prefix tree as illustrated in \cref{fig:prefix_horizon}, where counts at nodes track subwords to be generated and child nodes correspond to valid next subwords. See \cref{fig:decode} in \cref{sec:constrain:impl} for a concrete working example.

Conditional models use $p_\theta(y_{t} | \boldsymbol{y}_{<t}, \boldsymbol{x})$ to score the next token $y_t$ given previously generated tokens $\boldsymbol{y}_{<t}$. They additionally depend on the input $\boldsymbol{x}$, which helps track words to be generated and mitigate the ambiguity of selecting $y_t$. In contrast, unconditional models \cite{od:rush} with probability $p_\theta(y_{t} | \boldsymbol{y}_{<t})$ only depend on local information $y_t$ and $\boldsymbol{y}_{<t}$, which can lead to high ambiguity of selecting $y_t$ and attract beam search with small beams to local minimums. In \cref{sec:wo:conditional}, we analyze the difference between conditional and unconditional models with a fair comparison.

To instantiate $p_\theta(\boldsymbol{y} | \boldsymbol{x})$, we use a Transformer \cite{attention} consisting of both encoder and decoder pre-trained with BART \cite{bart}. Transformers use self-attention, a permutation invariant layer, to model contextual representations for input tokens. \citet{attention} add distinct position embeddings onto input token embeddings at different positions to make self-attention sensitive to input orders. BART pre-trains Transformers to reconstruct corrupted input sequences. As BART assumes sequential inputs, we convert the multiset input $\boldsymbol{x}$ into a pseudo-sequence by assigning an arbitrary permutation to the input words, following \citet{commongen}; see \cref{fig:diagram} for an illustration. Although subword orders within each word are informative, Transformers (and BART) may be sensitive to the arbitrary word permutation in the input. We analyze the permutation sensitivity of our models in \cref{sec:wo:perm}.

\subsection{Settings and Implementations}
\label{sec:wo:settings}
Following previous work \cite{od:hasler,od:tao}, we use PTB\footnote{We follow the \href{https://catalog.ldc.upenn.edu/license/ldc-non-members-agreement.pdf}{LDC User Agreement for Non-Members} license.} sections 2-21 (39,832 sentences) for training, section 22 (1,700 sentences) for development, and section 23 (2,416 sentences) for test. Punctuation escapes of PTB are reverted\footnote{We replace ``-LCB-''and ``-LRB-'' with ``('',  ``-RCB-'' and ``-RRB-'' with ``)'', while removing all ``\textbackslash''} to match the vocabulary of BART. We randomly shuffle words of each output sentence to create the input and perform BPE segmentation \cite{bpe} for both input and output. BLEU \cite{bleu} are reported as the performance metric following \citet{od:rush}. Our implementation is based on fairseq\footnote{\url{https://github.com/pytorch/fairseq}} \cite{fairseq}.

We train a Transformer from scratch (denoted \texttt{RAND}) as the baseline and compared it to the finetuned BART base (denoted \texttt{BART}) to estimate gains from BART pre-training. Hyperparameters are tuned separately since different models have different optimums. We find vocabulary size 8000 optimal for \texttt{RAND}. Both models share identical architecture with 6-layer encoder and decoder. \texttt{RAND} (35 million parameters) has smaller hidden size 512 and feed-forward hidden size 1024 compared to 756 and 3072 of \texttt{BART} (140 million parameters). They both need heavy regularization: $\beta=0.3$ for label smoothing \cite{ls}, $p=0.3$ for dropout \cite{dropout}, and $\alpha=1$ for R-drop \cite{rdrop}. Both models are trained using Adam \cite{adam}. We use 100 samples per batch, 4000 warm-up steps and learning rate 5e-4 for \texttt{RAND}; 20 samples per batch, 1000 warm-up steps and learning rate 1e-4 for \texttt{BART}. Learning rate decays with the inverse square root of training steps. We train the model till the development loss stops improving and average the last 5 checkpoints saved per 1000 training steps. We use beam size 64 to search on a constrained output space without additional specification. For unconstrained output space, we use beam size 64 and length normalization \cite{lengthnorm}.

\begin{table}[t]
    \centering
    \begin{tabular}{ l | c  c  c }
        Settings     & \texttt{B=5}   & \texttt{B=64}         & \texttt{B=512}        \\
        \midrule
        \multicolumn{4}{l}{\textit{Unconstrained}}                                          \\
        \texttt{AttnM}     & 34.89          & \multicolumn{1}{c}{/} & \multicolumn{1}{c}{/} \\
        \texttt{RAND}-ours & 38.53          & 38.95                 & 39.04                 \\
        \texttt{BART}-ours & \textbf{54.29} & \textbf{54.86}        & \textbf{54.84}        \\
        \midrule
        \multicolumn{4}{l}{\textit{Constrained}}                                            \\
        \texttt{Ngram}     & 23.3\hspace{0.5em}           & \multicolumn{1}{c}{/} & 35.7\hspace{0.5em}                  \\
        \texttt{RNNLM}     & 24.5\hspace{0.5em}           & 34.6\hspace{0.5em}                  & 38.6\hspace{0.5em}                  \\
        \texttt{Bag}       & 33.4\hspace{0.5em}           & 36.2\hspace{0.5em}                  & 37.1\hspace{0.5em}                  \\
        \texttt{RAND}-ours & 38.97          & 39.52                 & 39.59                 \\
        \texttt{BART}-ours & \textbf{54.70} & \textbf{56.38}        & \textbf{56.63}        \\
    \end{tabular}
    \caption{Test BLEU for word ordering on PTB. \texttt{B=X} denotes \emph{standard} beam search with beam size \texttt{X}. \texttt{Attn}, \texttt{Bag} and our models are all  conditional models while \texttt{Ngram} and \texttt{RNNLM} are unconditional ones.}
    \label{table:main_result}
\end{table}

\subsection{Word Ordering Results}
\label{sec:wo:results}
We compare our results with previous work under similar settings: for conditional models, bag2seq (\texttt{Bag}; \citealt{od:hasler}) and AttM (\texttt{AttnM}; \citealt{od:tao}) are included; for unconditional models, we include N-gram language models (\texttt{Ngram}) and RNNLM (\texttt{RNNLM}; \citealt{od:rush}) reproduced by \citet{od:hasler}.  Except for \texttt{AttnM}, all models use a constrained output space. We do not consider heuristically tailored beam search \cite{od:rush,od:hasler} and focus on standard sequence-to-sequence modeling. Different from existing studies, we use BPE segmentation for all our settings.

As shown in \cref{table:main_result}, our baseline \texttt{RAND} outperforms previous best results with unconstrained (38.53 of \texttt{RAND} compared to 34.89 of \texttt{AttnM} with \texttt{B=5}) and constrained (39.59 of \texttt{RAND} compared to 38.6 of \texttt{RNNLM} with \texttt{B=512}) output space, showing the effectiveness of sequence-to-sequence modeling with Transformer. Compared to \texttt{RAND}, \texttt{BART} brings substantial improvements under different settings, with up to 17.04 BLEU using \texttt{B=512} and constrained output space, which demonstrates the effectiveness of BART pre-training for word ordering, setting the stage for later analyses.
\begin{figure}[t]
    \centering
    \includegraphics[scale=0.48]{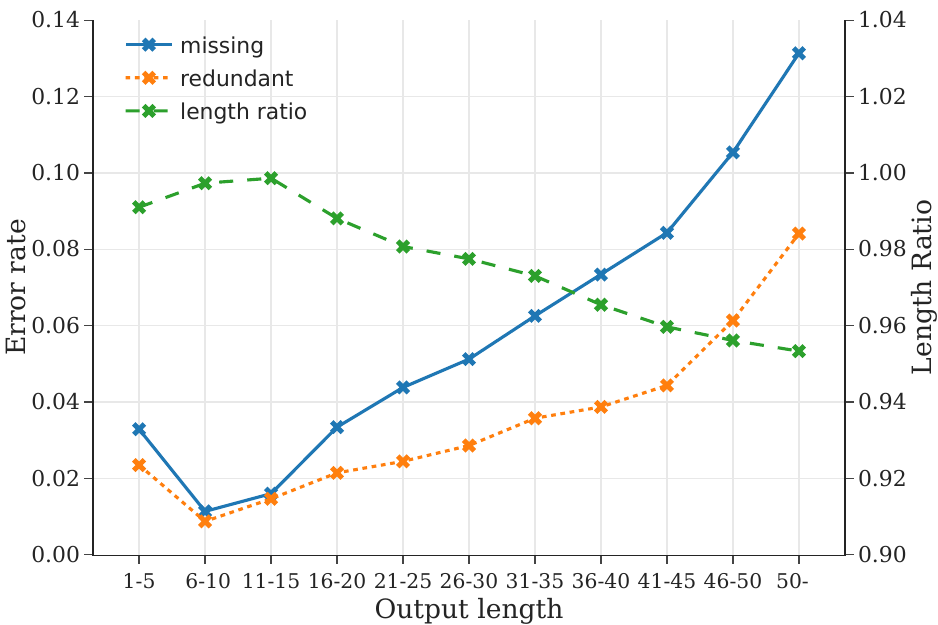}
    \caption{Test lexical errors of \texttt{RAND} with beam size 5 using unconstrained output space. We show the proportions of missing/redundant words and the output length ratios for test instances binned with output lengths. For the complete test set, the rates of missing and redundant words are 5.74\% and 3.3\%, respectively. The output length ratio is 0.981. See \cref{fig:constrain:full} in \cref{sec:unconstrain:error} for the results of different settings and how the metrics are computed.}
    \label{fig:constrain}
\end{figure}

\subsection{Errors of Unconstrained Output Space}
\label{sec:wo:constrain}
We can readily examine the output lexical errors by comparing the input and output bag of words. As shown in \cref{fig:constrain}, beam search with unconstrained output space tends to miss input words rather than generate redundant words. The tendency becomes prominent when increasing the output length, accompanied by a slight drop in the output length ratio. These lexical errors explain the consistent performance drop compared to constrained output space in \cref{table:main_result}. See \cref{sec:unconstrain:error} for additional results. The related coverage issue for sequence-to-sequence models has been studied in machine translation \cite{coverage,coveragemi}. However, different from word ordering, an error-prone source-target alignment procedure is required to estimate the output bag of words.

\subsection{Effects of Conditional Modeling}
\label{sec:wo:conditional}
We argued in \cref{sec:wo:modeling} that conditional modeling is less ambiguous when selecting the next token $y_t$, avoiding local minimum during beam search and thus performing well with small beams. Such a hypothesis is suggested by previous results included in \cref{table:main_result}: the conditional model \texttt{Bag} substantially outperforms (+8.9) the unconditional \texttt{RNNLM} with \texttt{B=5}; unconditional models require large beams to perform well \cite{od:rush}. We verify these observations with a \emph{fair} comparison. Concretely, we feed a null token as the input to simulate unconditional modeling with sequence-to-sequence models\footnote{The simulation leads to $P(\boldsymbol{x}=\mathrm{null}) = 1$. Thus we have $p_\theta(\boldsymbol{y}|\boldsymbol{x}=\mathrm{null}) = p_\theta(\boldsymbol{y})$} and train the models with the same settings in \cref{sec:wo:settings}.
\begin{figure}[t]
    \raggedright
    \includegraphics[scale=0.46]{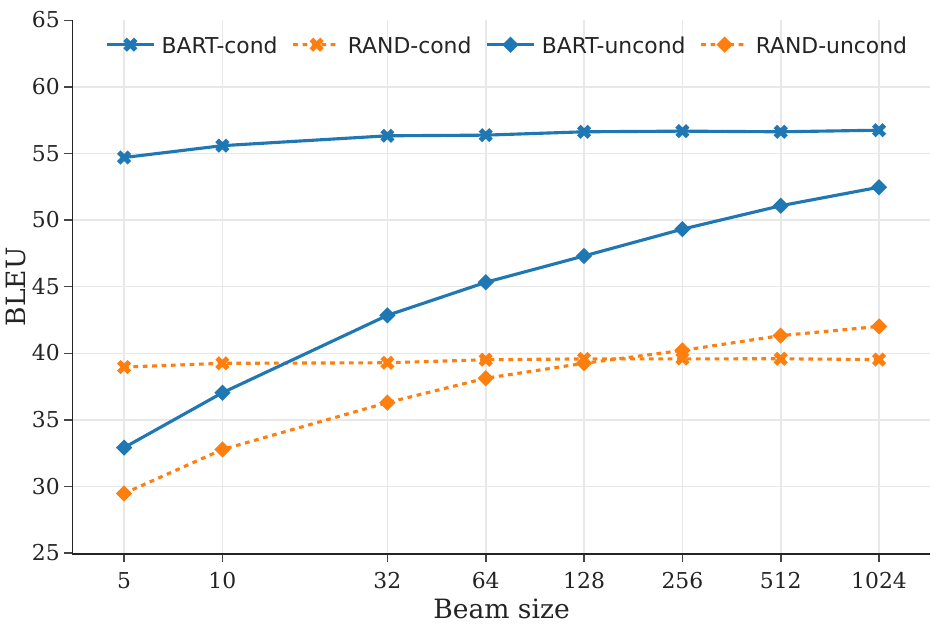}
    \caption{Test BLEU of conditional (\texttt{cond}) and unconditional (\texttt{uncond}) modeling with \texttt{RAND} and \texttt{BART}, using different beam size. A null token is fed to the encoder to simulate unconditional modeling.}
    \label{fig:unconditioal}
\end{figure}

Results are shown in \cref{fig:unconditioal}. With small beams, conditional models substantially outperform unconditional models. Unconditional models heavily rely on large beams to perform well. In contrast, small beams perform on par with large beams for conditional models. These results verify our hypothesis. Interestingly, as the beam size further increases, \texttt{RAND-uncond} slightly outperforms \texttt{RAND-cond}, showing that a larger candidate space can address ambiguities from local modeling $p_\theta(y_{t} | \boldsymbol{y}_{<t})$ to some extent, at the expense of extra computation and memory overhead.
\begin{figure*}[t]
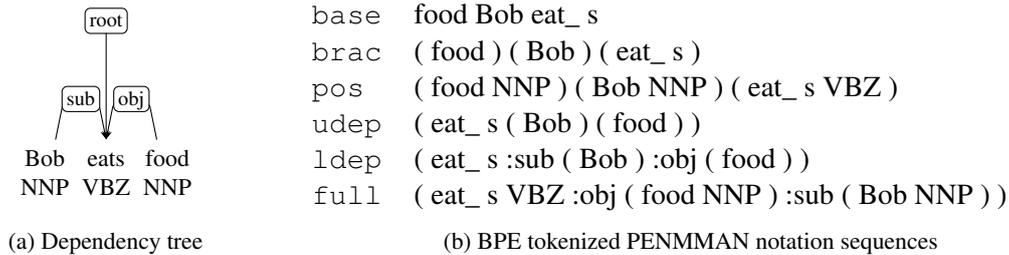

    \centering
    \begin{subfigure}[b]{.3\textwidth}
        \centering
        \begin{dependency}
            \begin{deptext}[font=\small]
                Bob \& eats \& food \\
                NNP \& VBZ \& NNP \\
            \end{deptext}
            \deproot[edge unit distance=2.6ex]{2}{root}
            \depedge[edge unit distance=3ex]{1}{2}{sub}
            \depedge[edge unit distance=3ex]{3}{2}{obj}
        \end{dependency}
        % \vspace{-0.6\baselineskip}
        \caption{Dependency tree}
        % \vspace{+0.5\baselineskip}
    \end{subfigure}
    \begin{subfigure}[b]{.65\textwidth}
        \begin{tabular}{l l}
            \texttt{base} & food Bob eat\_ s                                   \\
            \texttt{brac} & ( food ) ( Bob ) ( eat\_ s )                       \\
            \texttt{pos}  & ( food NNP ) ( Bob NNP ) ( eat\_ s VBZ )           \\
            \texttt{udep} & ( eat\_ s ( Bob ) ( food ) )                       \\
            \texttt{ldep} & ( eat\_ s :sub ( Bob ) :obj ( food ) )             \\
            \texttt{full} & ( eat\_ s VBZ :obj ( food NNP ) :sub ( Bob NNP ) ) \\
        \end{tabular}
        \caption{BPE tokenized PENMMAN notation sequences}
    \end{subfigure}
    \caption{An example dependency tree and its BPE tokenized PENMAN notation sequences with different features.}
    \label{fig:penman}
\end{figure*}
\subsection{Effects of Input Permutations}
\label{sec:wo:perm}

We empirically investigate the permutation sensitivity of our models for word ordering. The sensitivity is estimated with BLEU from 10 different development sets, each with distinct input \emph{word} permutations. We compare our models to several controlled settings. The first is data augmentation: for each training instance, we created augmented samples with the same target output but different input word permutations, denoted \texttt{aug}. The second is a Transformer without encoder position embeddings, which is invariant to input \emph{subword} permutations, denoted \texttt{npos}. To examine the importance of subword sequences, we also train \texttt{RAND} and \texttt{BART} with input subwords shuffled, denoted \texttt{shuf}. All models are trained with the same settings in \cref{sec:wo:settings}.

Results are shown in \cref{table:order}. Both \texttt{RAND} and \texttt{BART} with the baseline setting \texttt{base} are relatively insensitive to different input word permutations (with standard deviation 0.133 and 0.185), compared to the controlled setting \texttt{npos} (with standard deviation 0.05).\footnote{The quantization error of float arithmetic is sensitive to the order of operands.} Data augmentations \texttt{aug2}-\texttt{aug8} marginally improves the mean BLEU compared to \texttt{base}, but no consistent decrease of the standard deviation is observed. See \cref{sec:unconstrain:perm} for similar results with unconstrained output space. Surprisingly, with constrained output space, the lost local subword orders in \texttt{npos} and \texttt{shuf} has little impact on the performance, in contrast to the findings of \citet{perm:clou}. Even marginal improvement for \texttt{BART} is observed (56.45 with \texttt{shuf} compared to 56.21 with \texttt{base}). However, with unconstrained output space, the loss of local subword orders results in a non-trivial drop in performance. See \cref{sec:unconstrain:perm} for detailed results.

\begin{table}[t]
    \centering
    \begin{tabular}{ c |  c  c }
        Settings        & \texttt{RAND} & \texttt{BART} \\
        \midrule
        \texttt{base} & 40.45 (0.133) & 56.21 (0.185) \\
        \texttt{aug2} & 40.86 (0.159) & 56.78 (0.110) \\
        \texttt{aug4} & 40.73 (0.088) & 56.97 (0.164) \\
        \texttt{aug6} & 40.73 (0.157) & 56.76 (0.180) \\
        \texttt{aug8} & 40.94 (0.104) & 56.91 (0.155) \\
        \midrule
        \texttt{npos} & 40.05 (0.050) & /             \\
        \texttt{shuf} & 39.58 (0.135) & 56.45 (0.133) \\
    \end{tabular}
    \caption{Permutation sensitivity measured by the mean and standard deviation (in the bracket) of BLEU on 10 development sets with distinct input word permutations. Results are obtained with constrained output space; for unconstrained output space see \cref{table:order:unconstrain} in \cref{sec:unconstrain:perm}. \texttt{base} denotes settings in \cref{sec:wo:settings}. \texttt{augx} is trained with \emph{additional} \texttt{x} augmented samples per training instance. \texttt{npos} removes encoder position embeddings. \texttt{shuf} is trained with shuffled input subwords.}
    \label{table:order}
\end{table}

\section{Understanding Why BART Helps}
\label{sec:why}
We empirically establish the intuition that knowledge in BART helps target tasks. Though numerous types of knowledge have been identified in PLMs by probing \cite{bertology}, they may not necessarily improve the target task \cite{probe:improve}. Thus a reliable explanation should identify the \emph{relevant} type of knowledge for word ordering. Candidate types of knowledge can be selected using a procedure akin to feature importance (\cref{sec:why:relevance}): we feed different types of knowledge as additional features and select the one bringing the most salient gain in word ordering. Their relevance should be further verified by a strong correlation (\cref{sec:why:probe}) between the probing performance and word ordering performance, as models can utilize unexpected shortcut rules \cite{shortcut} instead of distilling the intended knowledge provided in the features.\footnote{The nuance is related to the philosophical quest on what constitutes mental states. Feature importance follows behaviorism while probing is akin to functionalism. See \cite{functionalism} for discussions on behaviorism and functionalism.} Such a correlation is estimated using models with different amount of the knowledge. We finally probe for the nontrivial existence of the knowledge in BART (\cref{sec:why:probe}).

\subsection{Analysis with Feature Importance}
\label{sec:why:relevance}
We first select a candidate type of knowledge for our explanation. Based on empirical evidence \cite{od:liu} and linguistic theories \cite{dep}, we narrow our focus to syntactic dependencies. Nevertheless, different parts of a dependency tree can be the candidate: brackets around words (\texttt{brac}), part-of-speech tags (\texttt{pos}), dependency structure (\texttt{udep}), labeled dependency structure (\texttt{ldep}), and the full tree with labeled dependency structure and part-of-speech tags (\texttt{full}). Knowledge are injected into models by feeding it as additional input feature. The resulting performance gain compared to the baseline (\texttt{base}) with bag-of-words inputs indicates the importance of the feature \cite{ablation}, a \emph{surrogate} for the relevance of the type of knowledge.

Dependency trees are derived from the PTB following \cite{partial:yue}, with tags defined by \citet{malt}. We use the same data split as in \cref{sec:wo:settings}. Bag-of-words inputs with additional features are turned into PENMAN notation sequences \cite{penman} and fed to sequence-to-sequence Transformers as in \cref{sec:wo:modeling}; see \cref{fig:penman} for input examples. For tree-structured inputs, we shuffle the children of each head nodes before turning them into sequences. Dependency labels and part-of-speech tags are kept intact during BPE tokenization. We follow the same settings in \cref{sec:wo:settings} to train \texttt{RAND} and \texttt{BART} with additional input features.

Results are shown in \cref{table:linearization}. The additional features from different types of knowledge consistently improves word ordering, among which dependency structure (\texttt{udep}) brings the main performance gain (comparing \texttt{udep} to \texttt{base}, \texttt{RAND} is improved by 47.15 and \texttt{BART} by 34.37), suggesting the \emph{potential} relevance of the knowledge to word ordering. Further adding dependency labels and part-of-speech tags marginally helps (comparing \texttt{ldep} and \texttt{full} to \texttt{udep}, \texttt{RAND} and \texttt{BART} are improved by up to 2.79 and 1.20, respectively). Interestingly, although part-of-speech tags alone slightly help (comparing \texttt{pos} to \texttt{base}, \texttt{RAND} is improved by 2.04 and \texttt{BART} by 1.49), their benefits diminish given dependency structures (comparing \texttt{ldep} to \texttt{full}, \texttt{RAND} is improved by 0.03 while \texttt{BART} dropping 0.27), suggesting that dependency structure knowledge can subsume part-of-speech tags. Accordingly, we select knowledge  about dependency structure as our candidate for explanation.

\subsection{Analysis with Structural Probing}
\label{sec:why:probe}
To obtain a reliable explanation, we then establish the correlation between dependency structure knowledge in the model and word ordering performance, and verify the existence of the knowledge in BART. Both require examining dependency structure knowledge in models, which can be achieved by the structural probe \cite{probe:syntax}, a learned bilinear distance metric taking representations of a model as input features. It estimates the unlabeled dependency trees of sentences using minimum spanning trees. The resulting unlabeled attachment score (UAS) indicates the probing performance. We use the average UAS of representations from all decoder layers to indicate the dependency structure knowledge in the model. We omit encoder representations since matching unordered input words to ground truth dependencies is ambiguous. Decoder representations are obtained by feeding the ground-truth outputs to the model. For each output token $y_t$, one can choose representations that predict it ($\mathbf{h}(\boldsymbol{y}_{<t}, \boldsymbol{x})$) or from feeding it ($\mathbf{h}(\boldsymbol{y}_{<t+1}, \boldsymbol{x})$) as features. We use the former as it results in higher UAS in our preliminary experiments. Features of words are the average of their subword features.

We follow the default hyperparameters of \citet{probe:syntax}\footnote{We use the code provided by the authors in \url{https://github.com/john-hewitt/structural-probes}}, with a rank of 32, and train with the L1 loss for 30 epochs using 40 samples per batch. We use the derived dependency trees from \cref{sec:why:relevance} as our dataset and report the averaged UAS on the PTB test set. Since dependency structure knowledge can subsume part-of-speech tags as shown in \cref{sec:why:relevance}, feeding features of \texttt{base}, \texttt{pos} or \texttt{udep} to \texttt{RAND} and \texttt{BART} results in models with varied amounts of the knowledge. Their structural probing results are shown in \cref{table:uas}.

\begin{table}[t]
    \centering
    \begin{tabular}{ c | l l | r}
        Settings    & \texttt{RAND}                 & \texttt{BART}                 & \multicolumn{1}{c}{$\Delta_{\texttt{B}-\texttt{R}}$} \\
        \midrule
        \texttt{base} & 40.36                         & 56.14                         & 15.78                        \\
        \texttt{brac} & 40.58 \textsuperscript{+\hspace{0.5em}0.22}  & 56.64 \textsuperscript{+\hspace{0.5em}0.50}  & 16.06                        \\
        \texttt{pos}  & 42.40 \textsuperscript{+\hspace{0.5em}2.04}  & 57.63 \textsuperscript{+\hspace{0.5em}1.49}  & 15.23                        \\
        \texttt{udep} & 87.51 \textsuperscript{+47.15} & 90.51 \textsuperscript{+34.37} & 3.00                         \\
        \texttt{ldep} & 90.27 \textsuperscript{+49.91} & 91.70 \textsuperscript{+35.56} & 1.43                         \\
        \texttt{full} & 90.30 \textsuperscript{+49.94} & 91.43 \textsuperscript{+35.29} & 1.13                         \\
    \end{tabular}
    \caption{Development BLEU with different input syntactic features: \texttt{brac} for brackets around words, \texttt{pos} for part-of-speech tags, \texttt{udep} and \texttt{ldep} for unlabeled and labeled dependencies, and \texttt{full} for part-of-speech tags and labeled dependencies. Performance gains against bag-of-words inputs (\texttt{base}) are shown in the superscripts. The $\Delta_{\texttt{B}-\texttt{R}}$ column is the differences between \texttt{BART} and \texttt{RAND}.}
    \label{table:linearization}
\end{table}

The consistent probing performance gains on feeding additional features in \cref{table:uas} confirms that knowledge is indeed injected into the models by feature feeding, ruling out the possibility that models use shortcut rules (with \texttt{pos}, \texttt{RAND} is improved by 1.26 and \texttt{BART} by 0.87; with \texttt{udep}, \texttt{RAND} is improved by 10.17 and \texttt{BART} by 12.13). Jointly examining \cref{table:linearization} and \cref{table:uas}, we find that an increase in UAS always corresponds to improved BLEU. The Pearson's correlation coefficient of 0.8845 between BLEU and UAS further \emph{verifies} that dependency structure knowledge is relevant to word ordering across settings.

We finally compare the probing performance of BART initialized with pre-trained parameters (with UAS 53.06) to the agnostic setting using randomly initialized token embeddings (with 42.59 UAS).\footnote{Our preliminary experiment shows that random embeddings substantially outperform features from randomly initialized Transformer (with 30.51 UAS).} The performance gap of 10.47 indicates that a nontrivial amount of dependency structure knowledge exists in BART. The relevance to word ordering and the existence in BART make knowledge about syntactic dependency structure a reliable explanation for the utility of BART in word ordering.

\section{Extension to Partial Tree Linearization}
\label{sec:partial}
Our analysis in \cref{sec:why} can be readily extended to partial tree linearization \cite{partial:yue}, a generalized word ordering task provided with additional syntactic input features. Unlike settings in \cref{sec:why}, additional features can be arbitrary \emph{subsets} of part-of-speech tags and labeled dependency arcs. The task can be helpful for applications such as machine translation \cite{partial:mt}. Following \citet{partial:yue}, we use the same dependency trees in \cref{sec:why:relevance} and report BLEU on the PTB development set with different proportions of syntactic features.

Previous studies \cite{partial:yue,partial:ratish} use linear models with hand-crafted features. Each base noun phrase (BNP; noun phrases without decedent noun phrases) is regarded as a single word for computation efficiency. We adopt the same sequence-to-sequence models \texttt{RAND} and \texttt{BART} in \cref{sec:wo:settings} and report results with and without special treatment for BNPs. Similar to \cref{sec:why:relevance}, partial trees are turned into PENMAN sequences and fed to sequence-to-sequence Transformers. We use the same model for different proportions of input features. For each tree in the training set, we sample 0\%, 50\% and 100\% of part-of-speech tags and labeled dependency arcs, respectively, resulting in 9 training instances with different input features for the same ordered output sequence. To keep inputs consistent, we put brackets around words that have no additional features (see \texttt{brac} in \cref{fig:penman}).

\begin{table}[t]
    \centering
    \begin{tabular}{ c | l l | r}
        Settings   & \texttt{RAND}                 & \texttt{BART}                 & \multicolumn{1}{c}{$\Delta_{\texttt{B}-\texttt{R}}$} \\
        \midrule
        \texttt{base} & 44.69                         & 53.05                         & 8.36                         \\
        \texttt{pos}  & 45.95 \textsuperscript{+\hspace{0.5em}1.26}  & 53.83 \textsuperscript{+\hspace{0.5em}0.78}  & 7.88                         \\
        \texttt{udep} & 54.86 \textsuperscript{+10.17} & 65.18 \textsuperscript{+12.13} & 10.32                        \\
        % \hline
        % \texttt{init} & 30.51 & 53.06 \\
        % \texttt{hinit}  & 28.50 & 54.09 \\
        % \texttt{einit}  & 42.59 & 46.89\\
    \end{tabular}
    \caption{Averaged UAS of all decoder layers (including the embedding) on test set. \texttt{base}, \texttt{pos} and \texttt{udep} are the same as \cref{table:linearization}. Performance gains against \texttt{base} are shown in the superscript. The differences between \texttt{BART} and \texttt{RAND} are shown in the $\Delta_{\texttt{B}-\texttt{R}}$ column.}
    \label{table:uas}
    % \texttt{hinit}, \texttt{einit} and \texttt{init} for \texttt{RAND} at random initialization and \texttt{BART} with pre-trained parameters, respectively. \texttt{hinit} denotes average UAS of decoder layers except the embedding, and \texttt{einit} for UAS of the embedding layer.
\end{table}

Results are shown in \cref{table:partial}. For comparison, we include results of \citet{partial:ratish}, denoted \texttt{P16}, and \citet{partial:yue}, denoted \texttt{Z13}. We notice that treating BNPs as words substantially simplifies the task: the mean BLEU increase from 59.5 to 73.0 for \texttt{RAND} and from 73.7 to 82.5 for \texttt{BART}. \texttt{RAND} substantially outperforms the previous best result of \texttt{P16} by 6.8 mean BLEU. In addition, \texttt{BART} brings further improvements by 9.5 mean BLEU, giving a new best-reported result.

\begin{table*}[t]
    \centering
    \begin{tabular}{ l | c c c c c c c c c | c}
        Settings                         & (0, 0)        & (0.5, 0)      & (1, 0)        & (0, 0.5)      & (0.5, 0.5)    & (1, 0.5)      & (0, 1)        & (0.5, 1)      & (1, 1)        & Mean          \\
        \midrule
        \texttt{Z13}\textsuperscript{†}  & 42.9          & 43.4          & 44.7          & 50.5          & 51.4          & 52.2          & 73.3          & 74.7          & 76.3          & 56.6          \\
        \texttt{P16}\textsuperscript{†}  & 48.0          & 49.0          & 51.5          & 59.0          & 62.0          & 67.1          & 82.8          & 86.2          & 89.9          & 66.2          \\
        \texttt{RAND}\textsuperscript{†} & 58.8          & 59.5          & 59.7          & 72.0          & 72.6          & 72.6          & 87.1          & 87.1          & 87.3          & 73.0          \\
        \texttt{BART}\textsuperscript{†} & \textbf{71.7} & \textbf{71.9} & \textbf{72.3} & \textbf{83.2} & \textbf{83.1} & \textbf{83.2} & \textbf{92.1} & \textbf{92.4} & \textbf{92.2} & \textbf{82.5} \\
        \midrule
        \texttt{RAND}                    & 42.0          & 42.4          & 42.8          & 55.2          & 55.6          & 56.8          & 80.1          & 80.1          & 80.5          & 59.5          \\
        \texttt{BART}                    & \textbf{55.9} & \textbf{56.5} & \textbf{57.2} & \textbf{73.6} & \textbf{73.6} & \textbf{73.6} & \textbf{90.9} & \textbf{90.8} & \textbf{90.9} & \textbf{73.7} \\
    \end{tabular}
    \caption{Development BLEU with constrained output space for partial tree linearization with different proportions of (\texttt{pos}, \texttt{ldep}) input features. Settings with † treat a BNP as a single word.}
    \label{table:partial}
\end{table*}

\section{Conclusion}
We investigated the role of PLMs in word ordering using BART as an instance. Non-sequential inputs are turned into sequences and fed to sequence-to-sequence Transformers and BART for coherent outputs. We achieve the best-reported results on word ordering and partial tree linearization with BART. With Transformers and BART, we provide a systematic study on the effects of output space constraints, conditional modeling, and permutation sensitivity of inputs for word ordering. Our findings can shed light on related pre-trained models such as T5 \cite{T5} in related tasks such as CommonGen. Our analysis with feature importance and structural probing empirically identifies that knowledge about syntactic dependency structure reliably explains the utility of BART in word ordering. Such a procedure is general and can be readily used to explain why a given PLM helps a target task.

\section*{Acknowledgements}
We would like to thank anonymous reviewers for their insightful comments and suggestions to help improve the paper. We gratefully acknowledge funding from the National Natural Science Foundation of China (NSFC No.61976180). Yue Zhang is the corresponding author.

\bibliography{anthology,custom}
\bibliographystyle{acl_natbib}

\appendix
\begin{figure*}[t]
    \centering
    \includegraphics[scale=0.3]{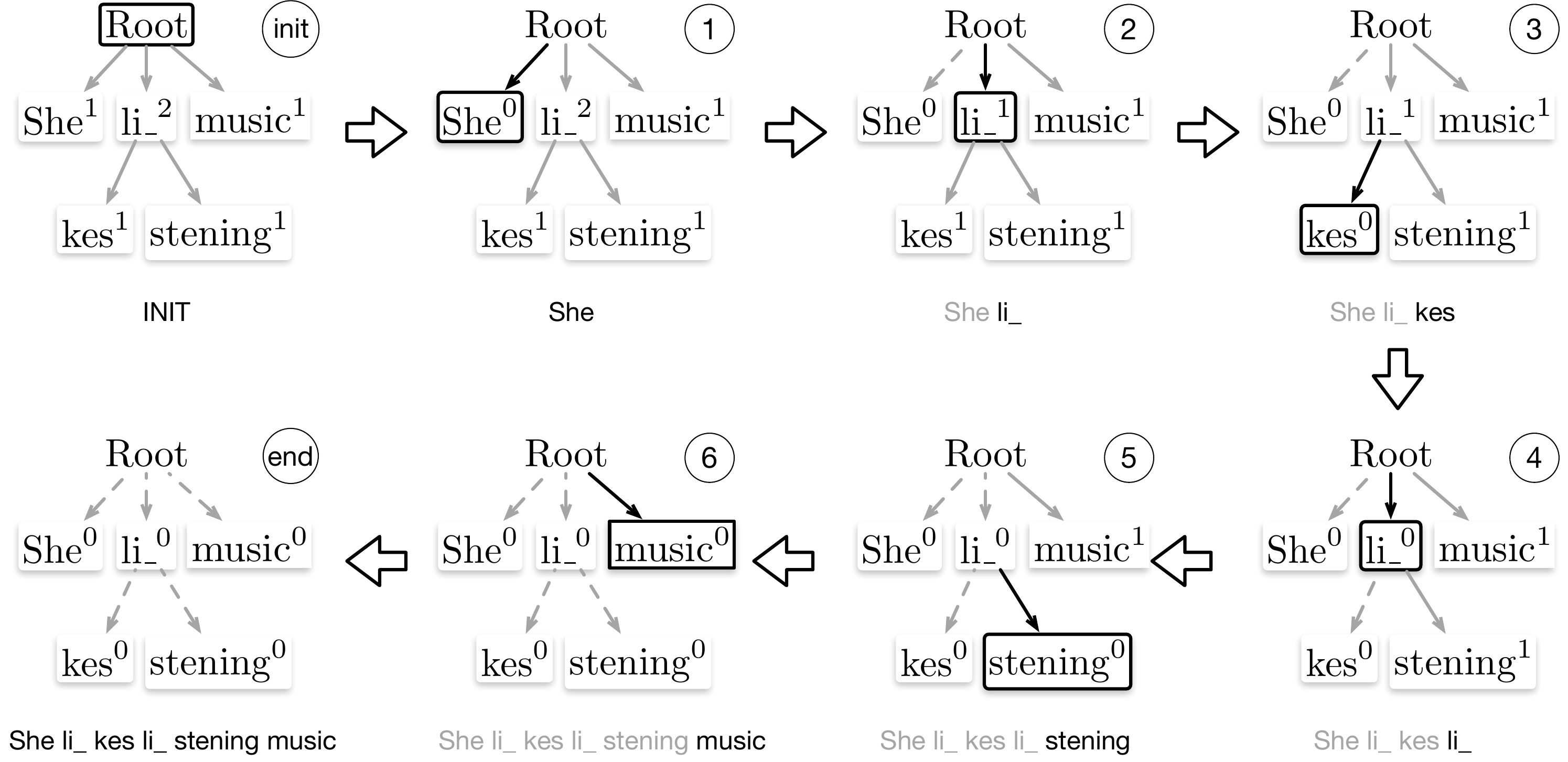}
    \caption{Illustration of how the constraint prefix tree in \cref{fig:prefix_horizon} of a path is updated during decoding. We omit the state of resetting the pointer to root for brevity. At initialization, the count of each node corresponds to how many times the subword appears in the input. At step 1, as ``She'' is a valid subword selected by beam search, we move the pointer the node of ``She'' and decrement its count by 1. In the following steps, ``She'' becomes invalid as its count becomes zero. After step 6, since counts in all children of the root become zeros, the path is marked finished.}

    \label{fig:decode}
\end{figure*}
\section{Beam Search with Output Constraints}
\label{sec:constrain:impl}
The standard beam search is modify by tracking the state of a candidate with a constraint prefix tree, which specifies valid next tokens at each decoding step. The update rules for the prefix tree are described in \cref{fig:prefix_horizon}. See \cref{fig:decode} for an illustration of how the constraint tree for a candidate in the beam is updated during decoding.

% \begin{figure}[t]
%     \centering
%     \includegraphics[scale=0.3]{figures/prefix.pdf}
%     \caption{Illustration of a constraint prefix tree defined by input subword sequences \{``She'', ``li\_ kes'', ``li\_ stening'', ``music''\}. Each node corresponds to a subword in the input, whose superscript is a counter for the times it appears in the input. The node with a bold outline is the current node pointed by the pointer.}
%     \label{fig:prefix}
% \end{figure}

\begin{table}[t]
    \centering
    \begin{tabular}{ c | c c c }
        Settings      & \texttt{RAND} & \texttt{BART} \\
        \midrule
        \texttt{base} & 39.39 (0.151) & 59.79 (0.158) \\
        \texttt{aug2} & 39.50 (0.118) & 60.00 (0.152) \\
        \texttt{aug4} & 40.08 (0.091) & 60.00 (0.167) \\
        \texttt{aug6} & 39.73 (0.053) & 59.86 (0.154) \\
        \texttt{aug8} & 40.02 (0.152) & 59.64 (0.170) \\
        \midrule
        \texttt{npos} & 36.98 (0.018) &               \\
        \texttt{shuf} & 36.52 (0.186) & 58.34 (0.124) \\
    \end{tabular}
    \caption{Permutation sensitivity measured by the mean and standard deviation (in the bracket) of BLEU on 10 development sets with distinct input word permutations. Results are obtained with unconstrained output space. Settings are the same as \cref{table:order}.}
    \label{table:order:unconstrain}
\end{table}

\section{Results of Unconstrained Output Space}
We include additional results with unconstrained output space to complement the discussion in \cref{sec:wo:constrain} and \cref{sec:wo:perm}.
\subsection{Lexical Errors}
\label{sec:unconstrain:error}
\begin{figure*}[t]
    \begin{subfigure}[b]{0.48\textwidth}
        \centering
        \includegraphics[scale=0.48]{figures/unconstrained_rand_beam5.pdf}
        \caption{\texttt{RAND,B=5}: missing 5.74\%, redundant 3.30\%, length ratio 0.981.}
    \end{subfigure}
    \hfill
    \begin{subfigure}[b]{0.48\textwidth}
        \centering
        \includegraphics[scale=0.48]{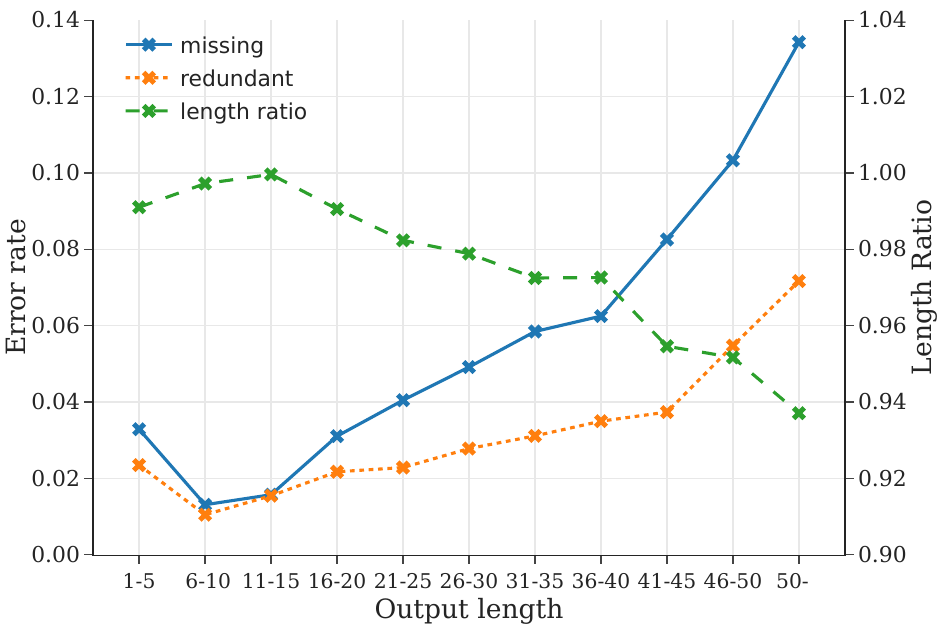}
        \caption{\texttt{RAND,B=512}: missing 5.42\%, redundant 3.02\%, length ratio 0.982.}
    \end{subfigure}
    \begin{subfigure}[b]{0.48\textwidth}
        \centering
        \includegraphics[scale=0.48]{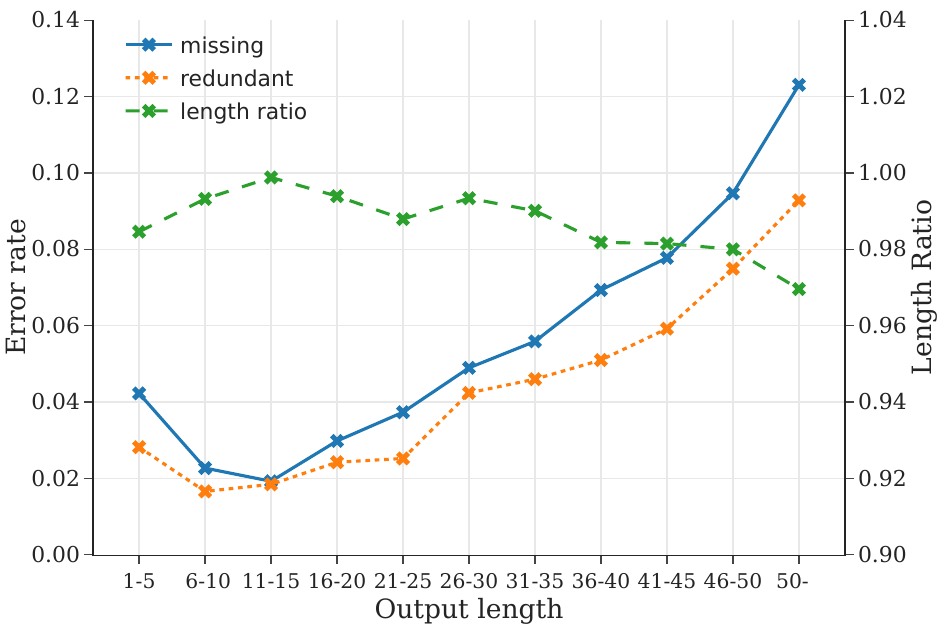}
        \caption{\texttt{BART,B=5}: missing 5.30\%, redundant 4.16\%, length ratio 0.990.}
    \end{subfigure}
    \hfill
    \begin{subfigure}[b]{0.48\textwidth}
        \centering
        \includegraphics[scale=0.48]{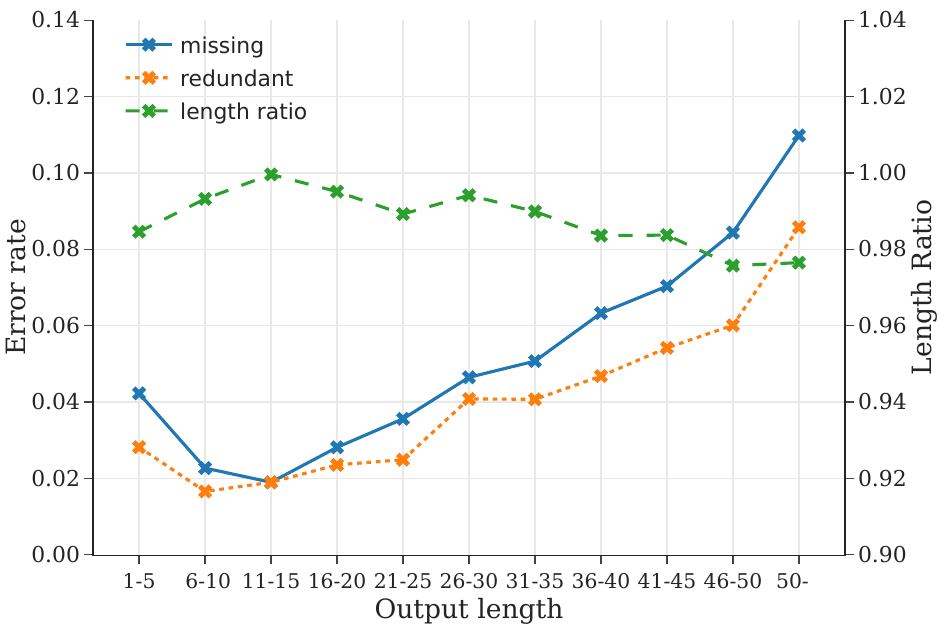}
        \caption{\texttt{BART,B=512}: missing 4.89\%, redundant 3.85\%, length ratio 0.991.}
    \end{subfigure}
    \caption{Test lexical errors with unconstrained output space similar to \cref{fig:constrain} for additional models and settings. We show the proportions of missing/redundant words and the output length ratios for test instances binned with output lengths. The model, beam size, and results on the complete test set are shown in the caption.}
    \label{fig:constrain:full}
\end{figure*}

In addition to \cref{fig:constrain} in \cref{sec:wo:constrain}, we present results for different models and beam sizes in \cref{fig:constrain:full}. Redundant (missing) words are all words in predicted (reference) output but not in reference (predicted) outputs. We normalize the count with number of words in all reference. Length ratio is ratio of predicted output length to reference output length.
\subsection{Permutation Sensitivity}
\label{sec:unconstrain:perm}
We replicate results of \cref{table:order} in \cref{sec:wo:perm} with unconstrained output space in \cref{table:order:unconstrain}. Similar to \cref{table:order}, data augmentation brings marginal improvement on mean BLEU and no consistent drop in standard deviation. Unlike \cref{table:order}, the loss of subword sequences results in a nontrivial drop in performance, especially for \texttt{RAND} (-2.87 from \texttt{base} to \texttt{shuf}). \texttt{BART} is less sensitive to subword shuffling (-1.45 from \texttt{base} to \texttt{shuf}).

\end{document}